\documentclass{article}
\usepackage{iclr2024_conference,times}

\usepackage{amsmath,amsfonts,bm}

\def\eqref#1{equation~\ref{#1}}
\def\1{\bm{1}}

\DeclareMathAlphabet{\mathsfit}{\encodingdefault}{\sfdefault}{m}{sl}
\SetMathAlphabet{\mathsfit}{bold}{\encodingdefault}{\sfdefault}{bx}{n}

\iclrfinalcopy

\usepackage[utf8]{inputenc} % allow utf-8 input
\usepackage[T1]{fontenc}    % use 8-bit T1 fonts
\usepackage{url}            % simple URL typesetting
\usepackage{booktabs}       % professional-quality tables
\usepackage{amsfonts}       % blackboard math symbols
\usepackage{nicefrac}       % compact symbols for 1/2, etc.
\usepackage{microtype}      % microtypography
\usepackage{xcolor}         % colors

\usepackage[ruled,vlined,linesnumbered, resetcount]{algorithm2e}
\usepackage{graphicx}

\usepackage{subcaption}
\usepackage{multirow}
\usepackage[normalem]{ulem}
\useunder{\uline}{\ul}{}
\usepackage{textcomp}
\usepackage{footnote}
\usepackage{makecell}
\usepackage{xspace}
\usepackage{enumitem}
\usepackage{mdwlist}
\usepackage{wrapfig,lipsum,booktabs}
\usepackage{amsthm}
\usepackage{pdfrender}

\usepackage[normalem]{ulem}
\useunder{\uline}{\ul}{}

\newcommand{\model}[0]{\textsc{PEM}\xspace}
\newcommand{\modelp}[0]{\textsc{PEMs}\xspace}
\newcommand{\taskpre}{{\mathcal{T}_{\text{pre}}}}
\def\datapre{{\mathcal{D}_{\text{pre}}}}
\newcommand{\randmask}[0]{\textsc{RandMask}\xspace}
\newcommand{\lastmask}[0]{\textsc{LastMask}\xspace}
\newcommand{\peakmask}[0]{\textsc{PeakMask}\xspace}
\newcommand{\seasondetect}[0]{\textsc{SeasonDetect}\xspace}
\newcommand{\autoformer}{\textsc{AutoFormer}\xspace}

\newcommand{\informer}{\textsc{Informer}\xspace}

\newcommand{\patchtst}{\textsc{PatchTST}\xspace}
\newcommand{\dlinear}{\textsc{DLinear}\xspace}
\newcommand{\timesnet}{\textsc{TimesNet}\xspace}
\newcommand{\micn}{\textsc{MICN}\xspace}
\newcommand{\funnel}{\textsc{FUNNEL}\xspace}
\newcommand{\epifnp}{\textsc{EpiFNP}\xspace}
\newcommand{\epideep}{\textsc{EpiDeep}\xspace}

\newcommand{\ebayes}{\textsc{EBayes}\xspace}

\title{PEMs: Pre-trained Epidemic Time-Series Models}

\author{%
  Harshavardhan Kamarthi, B. Aditya Prakash\\
  College of Computing\\
  Georgia Institute of Technology\\
  \texttt{\{harsha.pk,badityap\}@gatech.edu} \\
}
\begin{document}
\maketitle
\begin{abstract}
  Providing accurate and reliable predictions about the future of an epidemic is an important
problem for enabling informed public health decisions.
Recent works have shown that leveraging data-driven solutions that utilize advances in deep learning methods
to learn from past data of an epidemic often outperform traditional mechanistic models.
However, in many cases, the past data is sparse and
may not sufficiently capture the underlying dynamics.
While there exists a large amount of data from past epidemics,
leveraging prior knowledge from time-series data of other diseases is a non-trivial challenge.

Motivated by the success of pre-trained models in language and vision tasks, we tackle the problem
of pre-training epidemic time-series models to learn from multiple datasets from different diseases and epidemics.
We introduce Pre-trained Epidemic Time-Series Models (\modelp)
that learn from diverse time-series datasets of a variety of diseases by formulating pre-training as a set of self-supervised learning (SSL) tasks.
We tackle various important challenges specific to pre-training for epidemic time-series such as dealing with heterogeneous dynamics
and efficiently capturing useful patterns from multiple epidemic datasets by carefully designing the SSL tasks to learn important priors about the epidemic dynamics that can be leveraged for fine-tuning to multiple downstream tasks.
%We also propose using segments of time-series as input tokens to the transformer architecture of \model to better capture local temporal trends.
The resultant \model outperforms previous state-of-the-art methods in various downstream time-series tasks
across datasets of varying seasonal patterns, geography, and mechanism of contagion including the novel Covid-19 pandemic unseen in pre-trained data
with better efficiency using smaller fraction of datasets.
%Our work shows the efficacy of pre-training on cross-domain time-series datasets to improve downstream performance on a wide range of tasks.

\end{abstract}

%\keywords{Epidemic Forecasting, Self-supervised learning, Time Series Forecasting}

\section{Introduction}

Predicting the trends of an ongoing epidemic is an important public health problem that
influences real-time decision-making affecting millions of people.
Forecasting of time series of important epidemic indicators is a well-studied challenging problem~\citep{rodriguez2022data,chakraborty_what_2018}.
Availability of traditional as well as novel datasets such as
testing records, social media,  etc. that capture multiple facets of the epidemic
as well as advances in machine learning and deep learning in particular have enabled to build models that learn from these datasets and show promising results,
often outperforming traditional mechanistic methods~\citep{cramer2021evaluation,reich_collaborative_2019}.

\begin{figure}[h]
    \centering
    \includegraphics[width=.90\linewidth]{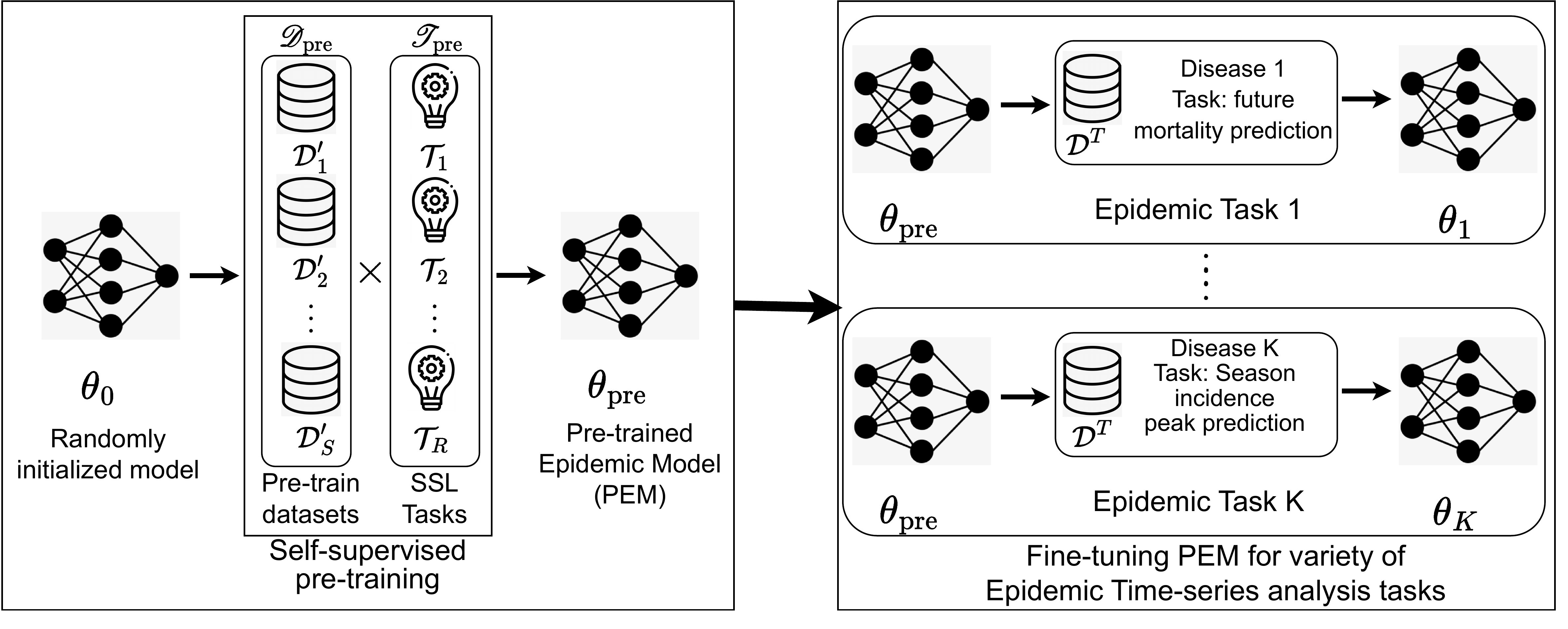}
    \caption{SSL based pre-training of \model over multiple tasks $\taskpre$
    and multiple pre-train datasets $\datapre$}
    \label{fig:start}
\end{figure}
Many public health and research initiatives collect data from various diseases over many decades
at various spatial granularities in different geographies.
Initiatives such as Project Tycho~\citep{van2018project} have aggregated these datasets that date back to the 1880s.
However, when building a data-driven model for an ongoing epidemic, we typically train the model \textit{only} using the
past data of the epidemic without leveraging useful knowledge information
and patterns from past epidemics.
In contrast, learning pipelines for language and vision use pre-trained models trained
on a much larger dataset~\citep{qiu2020pre,Du2022ASO,gunasekar2023textbooks}.
These pre-trained models are fine-tuned to the task of interest.
The benefits of the pre-trained models are two-fold.
First, the pre-trained weights are good initialization for faster and more effective training.
Moreover, pre-trained models learn useful underlying structures and patterns from larger pre-trained datasets such as
common syntactic and semantic knowledge in the case of language and the ability to recognize useful patterns  in the case of vision.
Initiating training from these pre-trained models usually results in faster training and better performance. %compared to starting model training from scratch.

Inspired by the success of pre-training in these fields, our goal is to leverage a large quantity of past epidemic time-series datasets
and build general pre-trained models that can be fine-tuned for multiple epidemic forecasting tasks on new unseen diseases.
However, there are challenges specific to time-series data that make pre-training non-trivial.
First, unlike text or images, epidemic time-series data is very heterogeneous.
Each dataset records different kinds of epidemic targets such as cases, mortality, hospitalization rates, etc., produced by different underlying dynamics from different geographical origins and can be affected by various levels of noise.
Moreover, while the size of the dataset potentially available for pre-training is typically larger than training data, it is still orders of magnitude smaller than in the case of text and images.
Indeed, previous state-of-art epidemic forecasting models have leveraged various methods such as sequence similarity~\cite{adhikari2019epideep,wang2020examining},
geographical relations~\cite{deng2020cola,wu2018deep}
using mechanistic priors~\cite{rodriguez2022einns,Gao2021STANSA,wang2021bridging}
ensembling~\cite{cramer2021evaluation,mcandrew2019adaptively}, etc. to overcome challenges related to
data sparsity and dealing with noise.
However, they can not effectively leverage heterogeneous datasets from other diseases and only use training data specific to the epidemic. Moreover, we do not have access to reliable auxiliary data such as mechanistic priors or spatial relations in most real-world epidemics.
%Further,  individual time-series datasets that compose the full set of pre-training data could be small and sparse in many cases.
Therefore, the pre-training methods we design for epidemic time-series datasets should be data-efficient and effective in learning useful general information about the heterogeneous epidemic datasets.
We tackle these challenges by viewing the problem of pre-training on multiple disease datasets via a self-supervised learning (SSL) framework.

While there have been multiple recent works on SSL for time-series data~\citep{yue2022ts2vec,tonekaboni2021unsupervised,eldele2021time,franceschi2019unsupervised},
these methods cannot be adapted to pre-training on multiple diseases.
Instead, they perform SSL on the training dataset specific to the task. % to improve the representation learning of the model.
Therefore, they cannot leverage useful epidemic priors during pre-training from multiple disease datasets.
Our work, in contrast, focuses on providing general-purpose 
pre-trained epidemic models
that can be fine-tuned to a wide range of epidemic analysis tasks for different diseases including novel and unseen epidemics by pre-training on datasets from various diseases of different  periodicity, sparsity,
 and the nature of underlying epidemic dynamics from the past.
We design three SSL tasks to leverage such multi-disease time-series datasets and learn pre-trained models
called \modelp (Pre-trained Epidemic Time-Series Models).
We also design our transformer-based architecture to enable us to better capture the temporal structure of time-series
and deal with datasets of varying magnitude and data distributions.
We show that PEMs can be fine-tuned to perform a wide variety of forecasting tasks for multiple diseases including novel diseases such as Covid-19 that are unseen in  pre-training.
Our contributions are summarized as follows:
    \noindent\textbf{(A) General pre-trained model for epidemic analysis tasks} We are the first to propose
    a single pre-trained model trained on multiple disease datasets that can be fine-tuned to a wide variety of downstream epidemic tasks concerning wide range of diseases.
    \noindent \textbf{(B) Self-supervised pre-training on cross-disease datasets}: We propose carefully designed SSL tasks that can learn from pre-train datasets of multiple diseases
           and efficiently capture important patterns of epidemic dynamics such as seasonality, and behavior around significant periods.
          We also note that, unlike text data, each time-stamp may not have sufficient semantic information.
          Similar to ~\cite{nie2022time}, we propose feeding segments of the time series as individual tokens to better capture local temporal trends.          
   \noindent \textbf{(C) State-of-the-art forecasting and peak prediction performance on multiple datasets}: We evaluate \model on disease datasets of different characteristics
          such as seasonality, geography, and mode of infection.
          We observed an 11-24\% improvement in performance over previous state-of-the-art baselines and SSL methods.
          \model also outperforms other methods on the task of forecasting mortality during the novel Covid-19 pandemic.
          \noindent \textbf{(D) Significant improvement in data and training efficiency and adaptability to novel epidemics}:
          \model also outperforms the baselines after significantly smaller training time and using smaller training data size compared to baselines.
          We also perform detailed ablation showing the importance of each of the proposed SSL tasks and modeling choices.
         % for \model's superior performance across a wide variety of epidemic time-series
          %analysis tasks.

%\input{Text/related_short.tex}
\section{Preliminaries}
\paragraph{Epidemic analysis tasks}

We focus on multiple tasks associated with forecasting on time-series of epidemic indicators of various diseases.
Informally, given the past time-series of an indicator of the epidemic such as case counts or mortality,
the goal is to predict specific characteristics that inform the future dynamics of the epidemic.
These targets include future values of the indicators (forecasting), predicting the time and magnitude of the peak or onset of the epidemic.
%These tasks can be modeled as regression or classification tasks.

Formally, let $\mathcal{D}^T$ be time-series data of epidemic indicators.
Let the time-series from time-stamp 1 to $T$ be denoted as $\mathbf{y}^{(1\dots T)} \in \mathbb{R}^{T}$.
The goal of an epidemic analysis task is to predict some useful property of the future values of the time-series $\mathbf{y}^{(T+1, \dots, T+K)}$.
For example, the task of forecasting involves predicting the values of $\mathbf{y}^{(T+1, \dots, T+K)}$.
Similarly, peak time prediction involves the prediction of time $t' = \arg\max_{t\in T+1 \dots T+K} y^{(t)}$
and peak intensity prediction involves estimating the value at the peak $\max_{t\in T+1 \dots T+K} y^{(t)}$.

\paragraph{Self-supervised pre-training for Epidemic analysis}

All previous methods only use datasets relevant to the specific epidemic to train or calibrate their models.
However, modelers typically also have access to time-series from other diseases collected in the past and
aggregated by initiatives like Project Tycho \cite{van2018project}.
Our work focuses on leveraging useful characteristics of epidemic data from these cross-disease data sources to provide
general pre-trained models that can better adapt to various epidemic analysis tasks.

Formally, along with the dataset $\mathcal{D}^t$ relevant to the specific epidemic of interest, we also have access to time-series datasets of other diseases from the past.
Let these set of datasets be denoted by $\datapre = \{\mathcal{D}'_1, \mathcal{D}'_2, \dots, \mathcal{D}'_S\}$.
We call $\datapre$ as \textit{pre-train datasets}.
Each of $\mathcal{D}'_j$ are univariate time-series datasets.
%The goal of pre-training is to leverage important patterns from datasets in $\datapre$ to initialize the weights of the PEMs that are then used to
%train for specific epidemic tasks.
We use the Self-supervised learning (SSL) framework to leverage $\datapre$ for pre-training.
We introduce a set of tasks $\taskpre=\{\mathcal{T}_i\}_{i=1}^R$.
At a high level, each task takes a time-series from $\datapre$, transforms it, and trains the model
to retrieve important properties of the input time-series.
These properties include
identifying seasons of certain segments of time-series, reconstructing important parts of the time-series, etc.

\paragraph{Problem Statement}
\textit{Given heterogeneous pre-train datasets $\datapre$ from multiple diseases, we aim to learn useful patterns, epidemic dynamics, and knowledge from $\datapre$ via SSL tasks $\taskpre$ such that the resultant pre-trained model can be fine-tuned to provide better performance on any epidemic analysis tasks for unknown diseases leveraging the generalizable patterns learned from $\datapre$.}

%We provide a detailed description of the SSL tasks $\taskpre$ we use in Section \ref{sec:ssl}.

%Formally, each task $\mathcal{T}_r$ consists of
%\begin{enumerate}
%    \item two mappings from a time-series $\mathbf{x}^{1\dots t}$ to a tuple $(f_r(\mathbf{x}^{1\dots t}), g_r(\mathbf{x}^{1\dots t}))$ where $f_r$ maps $\mathbf{x}^{1\dots t}$ to a transformed time-series and $g_r$  to a target that captures important property of $\mathbf{x}^{1\dots t}$
%    \item A loss function $\mathcal{L}_r$ that compares the predicted target of the pre-trained model to the actual target generated by $g_r$.
%\end{enumerate}

\section{Methodology}
\paragraph{Overview}

Our pipeline of leveraging pre-trained models for downstream epidemic tasks resembles that used in NLP and Vision problems.
Let $M(\theta_{pre})$ be the base PEM used for various epidemic analysis tasks and parameterized by $\theta_{pre}$.
%We assume that $M$  has a transformer encoder architecture since it can model sequential data and has shown success in.
$M$'s parameters are simultaneously pre-trained on each of the SSL tasks in $\taskpre$ on all the pre-train datasets.
This allows $M$ to learn from the underlying epidemic dynamics of multiple diseases without explicit supervision.
We then fine-tune the pre-trained $M(\theta_{pre})$ by first appending appropriate output layer $G(\theta_{last})$ to the model based on the task and
train all the $M$ and $G$ for the given downstream task.

\subsection{Segmented Transformer model}
\label{sec:segm}
\begin{wrapfigure}[14]{l}{.5\linewidth}
\vspace{-0.25in}
    \centering
    \includegraphics[width=.95\linewidth]{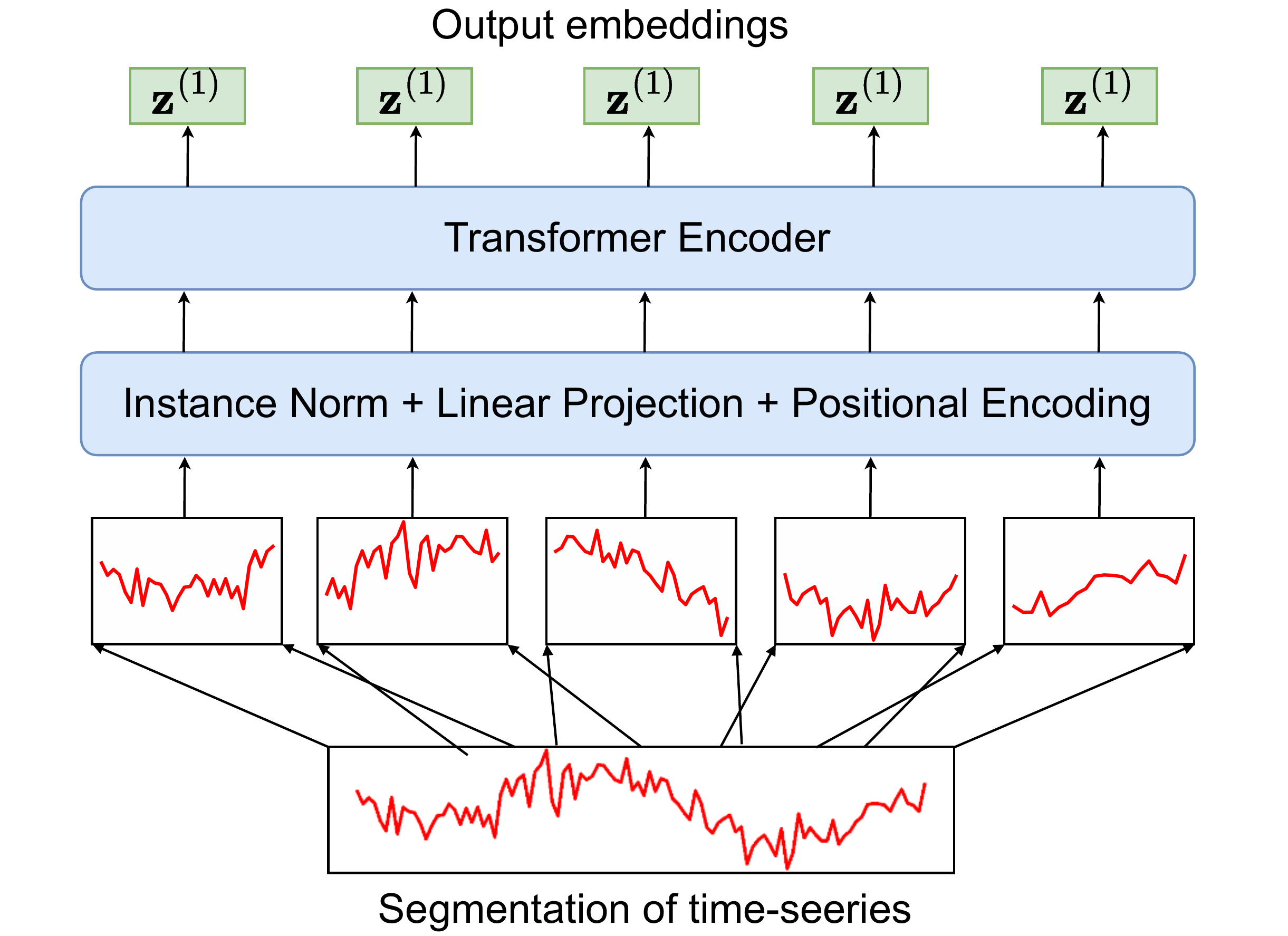}
    \caption{Segmented Transformer Architecture}
    \label{fig:trans}
\end{wrapfigure}
Transformers~\citep{vaswani2017attention} have been widely used in modeling sequential data, especially text data, and form the backbone architecture of large pre-traned models.
This is due to their ability to model long-range temporal relations as well as scale up to learn from large datasets during pre-training~\citep{dosovitskiy2020image,brown2020language}.
Recent works~\citep{zhou2021informer,chen2021autoformer} have shown the efficacy of transformers for
time-series forecasting in a wide range of domains.
Therefore, we design a transformer-based architecture for \model~s and modify it to better capture local temporal patterns. However, our framework can be easily extended to other neural sequential models like RNNs and convolutional networks.

%\hk{Don't cite~\cite{nie2022time} too much. Clarify in related works. Also same for other citations \cite{vaswani2017attention}}
%Due to their ability to model temporal patterns as well as scale to learn from large datasets for pre-training, we use a transformer architecture
%for PEMs.
Most previous works input features from each time step as a single token. However, unlike text data, each individual
time-stamp may not provide enough semantic meaning about temporal patterns of the time-series.
Therefore, similar to ~\cite{nie2022time}, we use segments of time-series, instead of individual time-stamps as input tokens.
We first segment the input time series $\mathbf{x}^{(1:T)} \in \mathbb{R}^{T}$ into uniform segments of size $P$ with stride length $S$
resulting in a sequence of length $L = \lfloor \frac{T-P}{S} \rfloor + 1$ denoted by $\mathbf{\hat{x}}^{(1:L)} \in \mathbb{R}^{P \times L}$.
Along with better temporal modeling,
segmenting the input can also enable the transformer model to efficiently process sequences of longer lengths since its inference speed scales
quadratically with sequence length.

For the first layer of the model, we project all the features of all the time stamps of each segment into an embedding and also inject positional information for each segment.
We first pass each segment $\mathbf{\hat{x}}^{(l)}$ of the sequence $\mathbf{\hat{x}}^{(1:L)}$ through a linear layer and add positional
embedding to its output:
\begin{equation}
    \mathbf{u}^{(l)} = \mathbf{W}_1\mathbf{\hat{x}}^{(l)} + \text{pos}(l)
\end{equation}
where $\mathbf{W}_i \in \mathbb{R}^{P \times D}$ and $\text{pos}(\cdot)$ is the positional encoding defined as
\begin{equation}
    \text{pos}(l) = \begin{cases}
        \sin(l/10^{5l/D})     & \text{if }l \text{ is even} \\
        \cos(l/10^{5(l-1)/D}) & \text{if }l \text{ is odd.}
    \end{cases}
\end{equation}
Positional encodings, therefore, help the model identify the absolute positions of the input segments.
The encodings $\{\mathbf{u}^{(l)}\}_{l=1}^{L}$ are fed into stacks of multi-head attention layers of the transformer similar to \cite{vaswani2017attention}
and we finally receive the output as a sequence of embeddings $\mathbf{z}^{(1:L)} \in \mathbb{R}^{D \times L}$.
For a specific SSL task or downstream fine-tuning tasks, we can append appropriate layers on top of $\mathbf{z}^{(1:L)}$ to generate outputs of desirable dimensions and properties.

\subsection{Self-Supervised learning}
\label{sec:ssl}
\begin{figure*}[!thb]
    \centering
    \includegraphics[width=\linewidth]{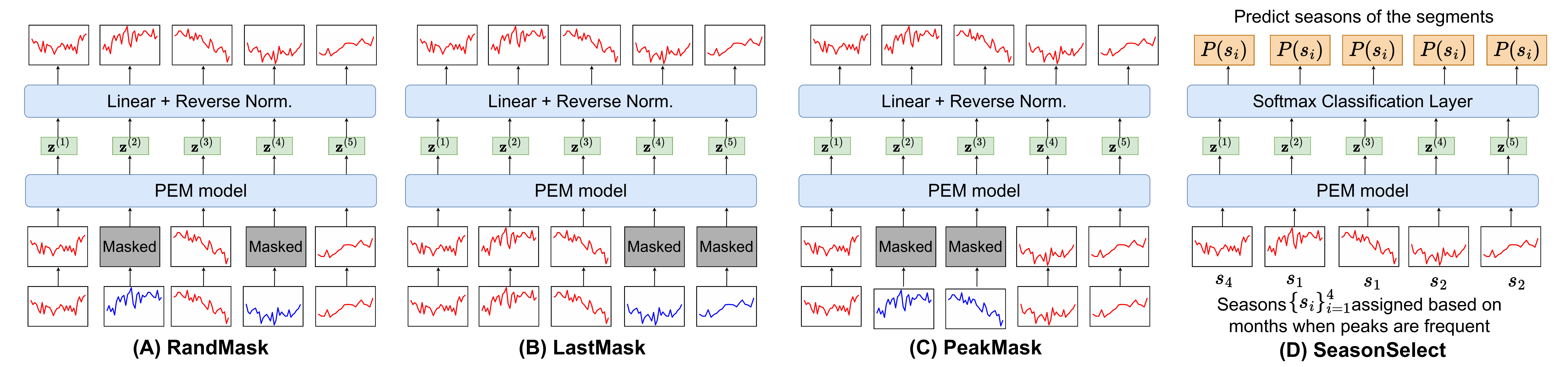}
    \caption{SSL task of $\taskpre$ are designed to efficiently capture
    important characteristics of epidemic dynamics by training on multiple disease datasets $\datapre$. %\hk{SeasonSelect top make season clear, bigger. Number A-D for all the tasks}
    }
    \label{fig:main}
\end{figure*}
The goal of pre-training is to provide the model with useful latent information about epidemic dynamics from
multiple disease datasets.
Unlike the case of text or images, we do not have a large pre-train dataset. Moreover,
the datasets for each disease is very heterogenous with each disease time-series measuring different epidemic indicator relevant to the disease and having large temporal
and spatial variance.
Therefore, we require our SSL tasks to be designed such that useful information from each of the disease dynamics is effectively captured.
We present the following four SSL tasks that capture various aspects of epidemic dynamics.

\paragraph{\randmask: Random Masking}
Epidemic forecasting models usually need to interpolate or extrapolate from input time-series to infer important characteristics of the epidemic.
Similar tasks have been proposed on individual text tokens for language models~\citep{devlin2018bert} to learn from large quantity of unlabeled text data.
%Inspired by the success of masked language modeling~\cite{devlin2018bert} for pre-training language models on a large amount of unlabeled text data,
%we propose a similar task for time series.
Random masking has also been explored for representation learning in previous works~\citep{zerveas2021transformer,nie2022time}.
However, they are applied on the same dataset as that used for training unlike our data and task-agnostic pre-training setup.
For our application, random masking allows the model to learn important patterns observed in typical epidemic time series.
Formally,
given the segment inputs $\mathbf{\hat{x}}^{(1:L)}$, we randomly mask $\gamma$ fraction of the segments with zero values and input to the model.
We apply a two-layer feed-forward network on top of each $\mathbf{z}^{(l)}$ to predict back all the values of segments for both masked and unmasked segments.
We use mean-squared error (MSE) loss to train for this task which we denote as \randmask($\gamma$).

\paragraph{\lastmask: Last segments masking}
Since most epidemic tasks involve predicting properties of the future of the epidemic including forecasting future values of the input time-series,
we propose another SSL task \lastmask that aims to re-construct the most recent segemnts of the input time-series.
Formally, \lastmask($\gamma$) is similar to \randmask$(\gamma)$, except instead of masking random $\gamma$ fraction of the segments, we mask only the last $\gamma$ fraction of the segments.
\lastmask, therefore, helps enable the model to predict the future dynamics of the epidemic by observing the past.

\paragraph{\peakmask: Masking around the peak}
Due to the chaotic nature of the epidemic curve around the peak, prediction of the shape of the curve around the peak is particularly hard.
Epidemic curves can peak due to many expected as well as unforeseen situations such as seasonal shifts due to weather, sudden outbreak, the introduction of new
variants of the pathogen strain or shifts in human behavior and policy changes.
\model needs to learn to anticipate and adapt to such scenarios.
Therefore, we introduce the \peakmask task which is another variant of \randmask.
For each input time series, we identify the time stamp with the maximum value.
We then mask all the segments that cover the maximum value's time-stamp and train the model to recover these segments.

\paragraph{\seasondetect: Seasonal detection}
Many diseases like influenza, chickenpox, Lyme disease, etc have seasonal patterns and peak at specific seasons.
Identifying the seasonal shifts in disease dynamics is important and valuable for many epidemic analysis tasks such as forecasting and peak detection.
We, therefore introduce \seasondetect, an SSL task for detecting the season of each time-series segment.
We first divide the 12 months of the year into 4 seasons: Season 1 (Dec-Feb), Season 2 (Mar-May), Season 3 (June-Aug), and Season 4 (Sept-Nov).
For each of the seasonal diseases $d$, in the pre-train dataset, we detect the season with the most peaks by calculating the month with the highest value in the time series
for all years in the dataset. We call this the peak season $s_1(d)$.
We then label the rest of the diseases relative to the peak season: $s_2(d)$ comes after $s_1(d)$, $s_3(d)$ comes after $s_2(d)$ and finally $s_4(d)$ comes after $s_3(d)$.
For example, in case of $d=\text{Influenza in US}$, $s_1(d)$ is assigned to months in Sept-Nov, $s_2(d)$ in Dec-Feb and so on.

The goal of the \seasondetect task is to identify the correct season among $\{s_i(d)\}_{i=1}^4$ for each of the input segments.
If a segment contains months from two seasons, it is assigned to the season which has the majority of the time steps (in case of a tie, the chronologically first season is assigned).
This problem is a segment classification problem.
Therefore, we apply a classification layer on top of each output embedding $\mathbf{z}^{(l)}$ that outputs the logits for the four seasons.
We use cross-entropy as the loss function for this task.

\subsection{Other Pre-training and fine-tuning details}
\label{sec:othertrain}

\paragraph{Dataset and Instance normalization}
The values of the time series of each dataset can vary widely.
For example, the ILI (influenza-related illness) indicators released by CDC are in the range of 0-10 whereas, for
other disease datasets such as Covid-19 or typhoid raw numbers on mortality or hospitalization are reported.
Therefore, as part of pre-processing we first normalize the time-series of each dataset of pre-train datasets independently.

Moreover, the data distribution and the magnitude of the time-series can vary across time for the same disease.
Therefore, we use reversible instance normalization~\citep{kim2021reversible} that performs instance normalization
on the input time series and reverses the normalization of the output of the model.

\paragraph{Multi-task training of all SSL tasks}
We pre-train our \model with all four SSL tasks on all pre-train datasets $\datapre$ together.
For each of the SSL tasks, we have a separate final layer but the parameters of the \model are shared across all tasks.
This is similar to hard parameter sharing used in multi-task learning~\citep{caruana1998multitask}.
For each batch, we randomly choose a dataset from $\datapre$ and randomly sample time-series from the dataset.
If the chosen dataset's disease is not seasonal, we skip training for \seasondetect for that batch.
This allows \model to learn from all the tasks, each of which imparts useful information from $\datapre$ which have varying utility for
different downstream tasks.

\paragraph{Linear-probing and fine-tuning}
\citet{Kumar2022FineTuningCD} showed that fine-tuning all the parameters of the pre-trained model for a specific downstream task can perform worse than just fine-tuning only the last layer (linear probing), especially for downstream tasks which are out-of-distribution to pre-trained data.
Since the downstream tasks in our case involve fine-tuning on novel diseases, \model suffers from this effect as well.
Therefore, based on their recommendation, we perform a two-stage fine-tuning process: we first perform linear probing followed by fine-tuning all the parameters.
\section{Experiments}
\label{sec:expt}

\paragraph{Datasets}
We leverage a large number of epidemic datasets aggregated by Project Tycho~\citep{van2018project}.
It has datasets from 1888 to 2021 for different diseases collected at state and city levels in the US.
While most of the datasets are collected on a weekly basis, many of these are very sparse
and have missing data.
We collect all the datasets that have time-series of length at least 10 to remove sparse data with short time-series.
We also use the weekly influenza data for US and Japan collected by CDC and NIID respectively.
Specifically, we use the aggregated and normalized counts of outpatients exhibiting
influenza-like symptoms released weekly by CDC\footnote{\url{https://gis.cdc.gov/grasp/fluview/fluportaldashboard.html}}.
For influenza in Japan, we use influenza-affected patient counts collected by NIID\footnote{\url{https://www.niid.go.jp/niid/en/idwr-e.html}}.
In total, we have 11 diseases: Hepatitis A, measles, mumps, pertussis, polio, rubella,
smallpox, diphtheria, influenza, typhoid and Cryptosporidiosis (Crypto.). 
%\hk{Emphasize different biology, disease dynamics,....}
We used all the data in Project Tycho till the year 1980 for each of these diseases for pre-training.
Since the influenza datasets are more recent and are collected from 2001 and 2010 for US and Japan respectively, we use influenza data up to 2012 from both countries for
pre-training. This set of disease datasets captures a rich variety of epidemic dynamics such as seasonality, mode of spread, underlying biology, etc.

%\aditya{you are not justifying why you are focusing on  crypto and typhoid}
\paragraph{Real-time Epidemic analysis tasks}
We evaluate the \model's performance on multiple diseases.
First, we evaluate forecasting for weekly influenza incidence in US and Japan, two geographically distinct locations, from 2013 to 2020.
We also perform forecasting on Cryptosporidiosia (Crypto.) from 2006 to 2012 and from 1980 to 1985 for typhoid.
Note that these are very diverse diseases with very different inherent dynamics: influenza is air-borne, and cryptosporidiosis and typhoid are water-borne.
While influenza and cryptosporidiosis are seasonal (Flu peaks during the winter while
cryptosporidiosis peaks during the summer),
typhoid has a stable and low incidence except for a few sudden outbreaks~\citep{matsubara2014funnel}.
We forecast the disease indicators at the national level for the entire country (US or Japan) for up to 4 weeks into the future.
Note that we use a real-time forecasting setup~\citep{reich_collaborative_2019,adhikari2019epideep} for training the model:
for forecasting up to four weeks from the current week, we use all the data till the current week to fine-tune the \model.
%\aditya{there are too many design choices for the setup. keep having citations so that people dont complain. e.g. week 40...}
We also perform peak week and intensity prediction for influenza similar to previous epidemic initiatives~\citep{reich_collaborative_2019}.
For each week in the season, we train the model to predict the week at which the peak occurs and the magnitude of the peak.
Note that the epidemic curve may have already peaked in the past when we train the model at the current week but the ground truth can only be known after
the full season.
Since the ground truth is a real number or an integer, we use the root mean squared error (RMSE) to evaluate all the tasks similar to~\citep{adhikari2019epideep,kamarthi2021doubt}.
(note that cross-entropy loss is used for peak week prediction due to discrete ground truth).

\paragraph{Baselines}
We compare \model with recent state-of-the-art (SOTA) models for both general time-series forecasting as well as models that are designed for epidemic analysis tasks.
The six recent state-of-the-art forecasting baselines are:
    \noindent$\bullet$ \informer(IF)~\citep{zhou2021informer} : Proposes an efficient sparse self-attention mechanism and a distillation mechanism to focus on the most important time-stamps.
     \noindent$\bullet$ \autoformer (AF)~\citep{chen2021autoformer}: Replaces self-attention with an auto-correlation mechanism to efficiently capture temporal dependencies.
     \noindent$\bullet$ \patchtst(PT)~\citep{nie2022time}: The backbone architecture used by \model that uniformly segments the input time-series into individual tokens of the transformer. Note that we do not perform any pre-training for PT baseline.
     \noindent$\bullet$ \dlinear(DL)~\citep{zeng2023transformers}: Uses linear layers instead of transformers for forecasting with comparable performance.
     \noindent$\bullet$ \timesnet(TN)~\citep{wu2022timesnet}: Uses a novel 2D inception block to model multiple temporal variations and periodicity in time-series.
     \noindent$\bullet$ \micn~\citep{wang2022micn}: Proposes multiple convolutional layers that capture patterns at multiple scales and merge them.
     %\noindent$\bullet$ \fedformer (FF)~\cite{zhou2022fedformer} : Borrows ideas from Fourier transform and seasonal-trend decomposition for efficient inference of long time-series.
     %\noindent$\bullet$ \pyraformer (PF)~\cite{liu2021pyraformer}: Introduces a pyramidal attention module for multi-scale relations with linear complexity.
We also compare against the following previous SOTA time-series epidemic forecasting models:
    %\noindent$\bullet$ \gaussianp~\cite{zimmer2020influenza}: Used Gaussian Process to provide seasonal flu predictions
    \noindent$\bullet$ \epideep (ED)~\citep{adhikari2019epideep}: Leverages similarity between current and historical time-series to provide interpretable forecasts
    \noindent$\bullet$ \epifnp (EF)~\citep{kamarthi2021doubt}: A previous state-of-art model for calibrated accurate forecasting extending Neural process framework~\cite{louizos2019functional}
          for sequential data.
    \noindent$\bullet$ \funnel (FL)~\citep{matsubara2014funnel} : Flexible Mechanistic model that can capture and forecast multiple epidemics
          by modeling useful characteristics like seasonality and sudden outbreaks
    \noindent$\bullet$ \ebayes (EB)~\citep{brooks2015flexible}: Used Empirical Bayes framework for flue forecasting and won previous FluSight competitions~\cite{reich_collaborative_2019}.

\section{Results}
\label{sec:results}

We evaluate \model through the following questions:
\textbf{Q1:} Does \model provide state-of-art performance in various epidemic analysis tasks?
\textbf{Q2:} Does pre-training enable faster and efficient training?
\textbf{Q3:} Is \model efficient in using less training data to provide consistently superior performance?
\textbf{Q4:} How does \model perform on diseases not available during pre-training?
\textbf{Q5:} How does \model compare against previous self-supervised time-series forecasting methods?
\textbf{Q6:} How do each of the SSL tasks and various other modeling choices affect the performance of \model?
We provide additional details on hyperparameters and training details in the Appendix \S \ref{sec:addntrain}.
and a link to code and datasets\footnote{Anonymized code link: \url{https://anonymous.4open.science/r/EmbedTS-3F5D/}}.

\paragraph{Forecasting performance (Q1)}
\label{sec:perf}
\begin{table}[h]
    \centering
    \vspace{-.15in}
    \caption{Weekly forecasting performance (RMSE) of \model and other top general forecasting and epidemic forecasting baselines. Top scores are in \textbf{bold} and the second best are {\ul underlined}. We observe the evaluation metrics scores to be statistically significantly better for \model using pair-wise t-test ($p\leq 0.05$) over 10 random runs.}
    \label{tab:forecast}
    \scalebox{0.8}{
        \begin{tabular}{c|c|cccccc|cccc|c}
                                    &            & \multicolumn{6}{c}{General-time Series}                & \multicolumn{4}{c}{Epi-Specific}   &                 \\ \hline
Datasets                            & Week ahead & AF   & IF            & PT   & DL   & TN   & MICN       & EF         & ED   & EB    & FUNNEL & PEM             \\ \hline
\multirow{5}{*}{\textbf{Flu-US}}    & 1          & 1.17 & 1.23          & 0.47 & 0.56 & 0.48 & 0.46       & {\ul 0.42} & 0.68 & 1.18  & 1.31   & \textbf{0.39}   \\
                                    & 2          & 1.28 & 1.37          & 0.83 & 0.72 & 0.51 & 0.52       & {\ul 0.48} & 0.73 & 1.26  & 1.33   & \textbf{0.42}   \\
                                    & 3          & 1.55 & 1.74          & 0.94 & 1.19 & 0.74 & {\ul 0.65} & 0.79       & 1.14 & 1.27  & 1.34   & \textbf{0.58}   \\
                                    & 4          & 1.64 & 2.11          & 1.16 & 1.25 & 0.97 & 0.81       & {\ul 0.78} & 1.81 & 1.34  & 1.37   & \textbf{0.61}   \\
                                    & Avg        & 1.41 & 1.61          & 0.85 & 0.93 & 0.68 & {\ul 0.61} & 0.62       & 1.09 & 1.26  & 1.34   & \textbf{0.50}   \\ \hline
\multirow{5}{*}{\textbf{Flu-Japan}} & 1          & 1139 & 1227          & 1205 & 944  & 922  & {\ul 934}  & 992        & 1186 & 1172  & 1388   & \textbf{831}    \\
                                    & 2          & 1572 & 1503          & 1517 & 1147 & 951  & {\ul 948}  & 1058       & 1395 & 1486  & 1694   & \textbf{894}    \\
                                    & 3          & 1676 & 1814          & 1667 & 1359 & 1189 & {\ul 1074} & 1179       & 1573 & 1858  & 1934   & \textbf{1035}   \\
                                    & 4          & 2044 & 1857          & 1918 & 1538 & 1488 & {\ul 1422} & 1572       & 1634 & 2297  & 2145   & \textbf{1069}   \\
                                    & Avg        & 1608 & 1600          & 1577 & 1247 & 1138 & {\ul 1095} & 1200       & 1447 & 1703  & 1790   & \textbf{957.25} \\ \hline
\multirow{5}{*}{\textbf{Crypto.}}   & 1          & 177  & 166           & 181  & 193  & 211  & 227        & {\ul 176}  & 211  & 205   & 229    & \textbf{147}    \\
                                    & 2          & 194  & 197           & 257  & 238  & 236  & 267        & {\ul 195}  & 259  & 381   & 415    & \textbf{176}    \\
                                    & 3          & 246  & 294           & 306  & 276  & 289  & 311        & {\ul 224}  & 327  & 496   & 614    & \textbf{205}    \\
                                    & 4          & 314  & 349           & 395  & 328  & 341  & 369        & {\ul 259}  & 411  & 642   & 665    & \textbf{239}    \\
                                    & Avg        & 233  & 252           & 285  & 259  & 269  & 294        & {\ul 214}  & 302  & 431   & 481    & \textbf{192}    \\ \hline
\multirow{5}{*}{\textbf{Typhoid}}   & 1          & 3.25 & \textbf{2.97} & 3.73 & 3.35 & 3.11 & {\ul 3.08} & 3.65       & 4.84 & 4.39  & 4.77   & \textbf{3.02}   \\
                                    & 2          & 4.19 & 3.93          & 5.78 & 4.06 & 3.76 & {\ul 3.66} & 3.97       & 5.11 & 5.33  & 5.13   & \textbf{3.38}   \\
                                    & 3          & 6.44 & 4.66          & 5.94 & 4.44 & 4.51 & {\ul 4.27} & 5.12       & 7.36 & 8.87  & 9.28   & \textbf{4.02}   \\
                                    & 4          & 6.98 & 5.19          & 6.94 & 5.13 & 5.22 & {\ul 4.42} & 5.93       & 8.95 & 11.98 & 13.22  & \textbf{4.61}   \\
                                    & Avg        & 5.22 & 4.19          & 5.60 & 4.25 & 4.15 & {\ul 3.86} & 4.67       & 6.57 & 7.64  & 8.10   & \textbf{3.76}  
\end{tabular}
    }
\end{table}
%\paragraph{Forecasting performance}
We perform real-time forecasting on four diseases discussed in \S \ref{sec:expt} for 1-4 weeks ahead and observe the performance in Table \ref{tab:forecast}.
We observe that \model's average forecasting performance outperforms both general time-series forecasting baselines as well as models specifically designed for epidemic
forecasting.
On average, \model provides 11-24\% improvement in RMSE.
The performance improvements are larger with longer forecast horizons.
%We observe a larger performance increase of around 25\% for four weeks ahead forecasting compared to around 1\% improvement for one week-ahead forecasts.
%\paragraph{Peak and onset prediction in Influenza}

We also perform prediction of time and intensity of onset and peak of the epidemic curves of seasonal diseases such as influenza  as done in past Flusight forecasting competitions organized by CDC~\citep{reich_collaborative_2019}.
The CDC defines the onset as the week at the past three consecutive weeks that have ILI above a baseline value defined by CDC.
For US dataset, these values are usually close to 2.2.
Since the Japanese data does not have onset baselines, we only perform peak time and intensity prediction.
For each of the baselines,  we append a classification head on top of the aggregated output embeddings
for peak and onset week prediction and train using cross-entropy loss.
For peak intensity prediction, we append a final layer that outputs a single scalar value and train using MSE loss.
The results are summarized in Table \ref{tab:onsetpeak}.
We observe that \model outperforms all baselines in most of the tasks and is comparable to the top-performing baseline for Peak intensity prediction in Influenza-Japan.
%On average we observe around 12\% improvement in Peak week prediction, 9\% improvement in Peak intensity prediction and 1\% improvement in onset week prediction.
This shows that \model can outperform the baselines in many epidemic analysis tasks across multiple diseases and geographical locations.

\begin{table}[h]
    \centering
    \caption{Peak and onset prediction performance of \model and baselines. Top performing scores are in \textbf{bold} and the second best is {\ul underlined}.}
    \label{tab:onsetpeak}
    \scalebox{0.9}{
    \begin{tabular}{c|c|cccccc|ccc|c}
                                    &                & \multicolumn{6}{c}{General-time Series} & \multicolumn{3}{c}{Epi-specific}       &               \\ \hline
Datasets                            & Task           & AF   & IF   & PT   & DL   & TN   & MICN & EF         & ED        & EB            & PEM           \\ \hline
\multirow{3}{*}{\textbf{Flu-US}}    & Peak week      & 6.37 & 7.33 & 8.38 & 8.91 & 7.32 & 7.15 & 6.92       & 6.3       & {\ul 5.22}    & \textbf{5.18} \\
                                    & Peak intensity & 0.94 & 0.93 & 1.13 & 1.14 & 1.27 & 1.03 & {\ul 0.85} & 0.97      & 1.05          & \textbf{0.72} \\
                                    & Onset week     & 6.49 & 8.33 & 9.17 & 9.34 & 7.41 & 7.55 & 7.26       & 6.11      & \textbf{5.28} & {\ul 6.11}    \\ \hline
\multirow{2}{*}{\textbf{Flu-Japan}} & Peak week      & 6.49 & 6.44 & 8.14 & 8.52 & 8.21 & 7.83 & 6.44       & 5.19      & {\ul 4.97}    & \textbf{4.72} \\
                                    & Peak intensity & 983  & 1066 & 1289 & 1187 & 1052 & 974  & 915        & {\ul 874} & 1046          & \textbf{864} 
\end{tabular}
}
\end{table}

\paragraph{Effect of pre-training on efficient fine-tuning (Q2)}
The time taken to train \textit{till convergence}  and memory requirements for \model is similar to the transformer-based baselines. \model takes 47-88\% of training time taken by state-of-art baselines to reach similar performance. This shows that pre-training enables both \textit{faster as well as more effective training}. See Appendix \S \ref{sec:timemem} for details. 

\paragraph{Data efficiency of \model (Q3)}
Pre-training enables large language models to adapt ot downstream tasks using small amount of data~\citep{brown2020language}.
We measured the performance of \model using various fractions of training data and observed that \model typically equals or outperforms best baselines using 60-80\% of training data (See Appendix Figure \ref{fig:perf_percent}).

\paragraph{Generalization to novel diseases (Q4)} 
In real-world scenarios we may encounter novel epidemic dynamics not seen in pre-train datasets such as outbreak of a novel disease or deploying the model to novel geographical location. Therefore, we measure the efficacy of \model when we remove the disease in downstream task from pre-training data.
 PEM provides state-of-art performance in most cases. We also show that \model can \textit{adapt to the dynamics of the novel Covid-19 pandemic}, which is not present during pre-training by providing 2-12\% better performance over previous state-of-art models. The detailed results are discussed in Appendix \S \ref{sec:adapt}.

\paragraph{Comparison with alternate SSL methods (Q5)}
\label{sec:sslresults}
\begin{table}[h]
\vspace{-.15in}
    \centering
    \caption{Comparison of \model with alternate SSL methods and pre-training with each of the SSL tasks in $\taskpre$ independently.}
    \label{tab:ssl}
    \scalebox{0.90}{
        \begin{tabular}{c|cccc|cc|cc}
                                                    & \multicolumn{4}{c}{Forecasting}                                                                                                              & \multicolumn{2}{c}{Peak week}                                          & \multicolumn{2}{c}{Peak intensity}                                     \\ \hline
Model             & \multicolumn{1}{l}{Flu-US} & \multicolumn{1}{l}{Flu-japan} & \multicolumn{1}{l}{Crypto.} & \multicolumn{1}{l}{Typhoid} & \multicolumn{1}{l}{Flu-US} & \multicolumn{1}{l}{Flu-japan} & \multicolumn{1}{l}{Flu-US} & \multicolumn{1}{l}{Flu-Japan} \\ \hline
SSL Methods       & \multicolumn{1}{l}{}             & \multicolumn{1}{l}{}                & \multicolumn{1}{l}{}                  & \multicolumn{1}{l}{}        & \multicolumn{1}{l}{}             & \multicolumn{1}{l}{}                & \multicolumn{1}{l}{}             & \multicolumn{1}{l}{}                \\ \hline
TS2Vec            & 1.85                             & 1175.3                              & 247.6                                 & 6.11                        & 7.33                             & 7.22                                & 0.95                             & 1198                                \\
TNC               & 1.22                             & {\ul 1059.6}                        & 317.4                                 & 7.92                        & 8.18                             & 6.91                                & 0.83                             & 1045                                \\
TS-TCC            & 1.94                             & 1344.6                              & 306.6                                 & 5.68                        & 7.94                             & 6.88                                & 1.05                             & 1079                                \\ \hline
Ablation variants & \multicolumn{1}{l}{}             & \multicolumn{1}{l}{}                & \multicolumn{1}{l}{}                  & \multicolumn{1}{l}{}        & \multicolumn{1}{l}{}             & \multicolumn{1}{l}{}                & \multicolumn{1}{l}{}             & \multicolumn{1}{l}{}                \\ \hline
No Pre-training   & 0.85                             & 1577.3                              & 285                                   & 5.61                        & 8.38                             & 8.14                                & 1.13                             & 1287                                \\
No Linear Probing & {\ul 0.51}                       & {\ul 979.5}                         & {\ul 211.5}                           & {\ul 4.06}                  & \textbf{5.12}                    & {\ul 4.88}                          & \textbf{0.68}                    & {\ul 882}                           \\
Only \randmask     & 0.85                             & 1473.5                              & 238.2                                 & 5.11                        & 9.22                             & 6.94                                & 0.98                             & 1055                                \\
Only \peakmask     & 0.63                             & 1244.7                              & 238.7                                 & 5.19                        & 5.82                             & 5.27                                & 0.82                             & 885                                 \\
Only \lastmask     & 0.72                             & 1129.5                              & 222.4                                 & 4.17                        & 6.11                             & 6.25                                & 0.96                             & 1074                                \\
Only SeasonSelect & 0.71                             & 1443.6                              & 227.3                                 & 5.29                        & 7.3                              & 7.19                                & 0.93                             & 917                                 \\\hline
\model               & \textbf{0.50}                    & \textbf{957.2}                      & \textbf{192}                          & \textbf{3.76}               & {\ul 5.18}                       & \textbf{4.72}                       & {\ul 0.72}                       & \textbf{864}                       

        \end{tabular}
    }
\end{table}
Using SSL to improve representation learning of time-series has been explored in
prior works. %~\citep{yue2022ts2vec,tonekaboni2021unsupervised,eldele2021time,nie2022time}.
However, the goal of these methods is narrow: they only aim to improve performance for a specific task by
performing SSL on the same training dataset.
Therefore, these methods may not efficiently learn from multiple disease datasets and 
generalize well to a wide range of downstream tasks.
%We tackle these issues of  dataset heterogeneity and learning from multiple small epidemic datasets by designing SSL methods (\S \ref{sec:ssl})
%that efficiently extract useful patterns related to epidemic analysis tasks
%such as modeling dynamics around the peaks, capturing seasonality, and learning to forecast future values.
We compare \model with these previous works:
TS2Vec~\citep{yue2022ts2vec}, TNC~\citep{tonekaboni2021unsupervised} and TCC~\citep{eldele2021time}.
We use these methods to train on all of pre-train
datasets $\datapre$ and fine-tune the models for each of the tasks by appending the appropriate output layer.
We also measure the impact of each of the SSL tasks $\taskpre$ on all the benchmarks.
Finally, we also examine the impact of linear probing and pre-training as a whole.

The results are summarized in Table \ref{tab:ssl}.
\model outperforms the alternate SSL methods that fail to even beat top baselines in most cases.
We see over 15\% improvement in all forecasting tasks, over 31\% improvement in peak week prediction, and 27\% improvement in peak intensity prediction.
%This illustrates the importance of the careful design of our SSL tasks for epidemic analysis tasks compared to other SSL methods that are designed for general time-series
%and may not adapt well to pre-train datasets from multiple disease sources.
With regard to the impact of each of the tasks in $\taskpre$, we observe that \model pre-trained with all of the tasks performs better
those pre-trained with any single SSL tasks.
For forecasting tasks, we find that \peakmask or \lastmask are the most important SSL methods.
%As expected, \peakmask is the most useful SSL task for peak week prediction tasks.
%Using only \randmask similar to~\cite{nie2022time}, in general, seems to be worse compared to using any other SSL task.
We also observe that two-step pre-training helps improve downstream performance and without any pre-training the model performance deteriorates to a large extent,
significantly underperforming many of the baselines.

\paragraph{Modeling choices and Hyperparameter sensitivity (Q6)} 
We also study the impact of various model design choices and the sensitivity of important hyperparameters. The detailed results of the study are presented in Appendix \S \ref{sec:ablation}.
We studied the impact of two important model design choices: segmentation (\S \ref{sec:segm}) and instance normalization (\S \ref{sec:othertrain}). We observed a 27-75\% decrease in performance by tokenizing individual time-steps instead of segments and an 8-31\% decrease in performance without instance normalization.
We studied the effect of SSL hyperparameters and segment size. The  hyperparameters of the segment size ($P=4$), $\gamma=0.2$ for \randmask and $\gamma=0.1$ for \lastmask generally perform the best if not close to best across multiple diseases.
%Moreover, the performance was not significantly affected by small changes to other hyperparameters.

\section{Conclusion}

We study the challenge of leveraging heterogeneous epidemic time-series data across multiple sources via pre-trained models 
that can be fine-tuned to outperform previous state-of-the-art models in diverse epidemic analysis tasks
concerning a wide range of diseases.
\model provided 11-24\% better forecasting performance and 9-12\% improvement in peak prediction outperforming
strong general time-series forecasting and epidemic-specific forecasting models.
We also showed the importance of specific modeling choices such as segmenting input time-series ~\citep{nie2022time} as individual tokens and more importantly, designing effective SSL tasks that can learn from cross-disease datasets
of multiple diseases. %by observing that applying prior SSL methods underperformed even some of the baselines that did not leverage any pre-training.
%Our ablation studies showed the effectiveness of our various modeling choices such as segmenting the input time-series as individual tokens as well as using each of the proposed SSL tasks in over.
Our work is the first to show the efficacy of pre-training on a wide array of unlabeled datasets from multiple unrelated sources as a viable
method to improve model performance across multiple applications.
While our backbone architecture is fairly straightforward, our SSL methods can be easily used to extend a wide range of model architectures that ingest time-series data.
Our work could also lead to further important research in the direction of general pre-trained models for time-series
similar to pre-trained models in language and vision domains.
%While we use disease datasets from a wide range of standard data sources collected by Project Tycho~\cite{van2018project}, CDC, and NIID, $\datapre$ can potentially be extended to include epidemic
%datasets from other regions or even other surveillance sources like
%social media~\cite{yang2015accurate}, weather~\cite{soebiyanto2010modeling}, etc.
Our approach can be extended to other applications such as health care, economics, and sales which typically have
a large number of datasets from multiple sources.
We can also easily extend to other time-series tasks such as classification and anomaly detection.

Our work is limited to time-series data whereas other multimodal data epidemic sources may also be available for diseases such as mobility networks, geographical relations, social media, etc. that cannot be directly integrated into \model.
Adapting \model to leverage these features of varying temporal scales~\citep{rodriguez2022data,ibrahim2021variational} and providing probabilistic forecasts with uncertainty quantification~\citep{xu2021conformal,kamarthi2021doubt} are
important research directions.
Due to our method's relevance to critical public health applications and decision-making, the potential misuse of our model can not be discounted.
Steps should be taken to alleviate problems such as disparities in the quality of data collected across regions, equity of prediction performance, etc.

\bibliography{references}
\bibliographystyle{iclr2024_conference}

\newpage
\appendix

{\Large \bf Appendix for PEMs: Pre-trained Epidemic Time-Series Models}

\section{Related works}
\paragraph{Neural models for time-series analysis}
Deep neural networks have been widely used in many time series forecasting
applications with great success.
DeepAR~\cite{salinas2020deepar} is a popular forecasting model that trains an auto-regressive recurrent network to predict the parameters of the forecast distributions. 
Other works including
deep Markov models \cite{krishnan2017structured} and deep state space models \cite{rangapuram2018deep, li2021learning, gu2021efficiently} explicitly model the transition and emission components with neural networks.
Recent works have also leveraged transformer-based models, which
have been widely used for language modeling, on
general time-series forecasting~\cite{oreshkin2019n}.
Other works have extended the transformer architecture
to improve efficiency and better capture long-term temporal
trends resulting in state-of-art performance in many
long-term forecasting benchmarks~\cite{zhou2021informer,chen2021autoformer,zhou2022fedformer,liu2021pyraformer}.
However, all these methods do not leverage pre-training.
They follow the typical supervised learning paradigm of
leveraging training data from past values of the same dataset to
forecast future values and do not leverage cross-domain heterogenous
datasets or aim to provide generalized models that can be used for a wide range of
heterogeneous tasks.

\paragraph{Self-supervised learning for time-series}
Recent works have shown the efficacy of
self-supervised representation learning for time-series for various classification
and forecasting tasks in wide range of applications such as
modeling behavioral datasets~\cite{merrill2022self,chowdhury2022tarnet},
power generation~\cite{zhang2019deep}, health care~\cite{zhang2022self}.
\citet{franceschi2019unsupervised} used triplet loss to discriminate segments of the same time-series from others.
TS-TCC used contrastive loss with different augmentations of time-series~\cite{eldele2021time}.
TNC~\cite{tonekaboni2021unsupervised} use the idea of leveraging neighborhood similarity for unsupervised learning of local distribution of temporal dynamics.
TS2Vec leveraged hierarchical contrastive loss across multiple scales of the time-series~\cite{yue2022ts2vec}.
However, all these methods apply SSL on the same dataset that is used for training and may not adapt well to using time-series multiple sources
such as time-series from multiple diseases.
Our work, in contrast, tackles the problem of learning general models
from a wide range of heterogenous datasets that can be fine-tuned
for a wide variety of tasks on multiple datasets that may not be
used during pre-training.
Therefore, we design SSL tasks that can adapt to multiple time-series datasets and capture useful underlying properties from
these datasets for superior performance on multiple downstream applications on various disease forecasting tasks.

\paragraph{Statistical models for epidemic forecasting}
Due to recent advances in machine learning and deep learning as well as the availability of datasets from various surveillance
sources, statistical and deep-learning-based models are increasingly used for epidemic forecasting tasks with great success~\cite{rodriguez2022data}.
Classical auto-regressive time-series models like ARIMA and its variants
have been adapted for disease forecasting~\cite{soebiyanto2010modeling,yang2015accurate}.
Other models use Bayesian generative approach~\cite{brooks2015flexible,brooks2018nonmechanistic} to provide probabilistic forecasts
and have been successful in past epidemic forecasting competitions like
Flusight~\cite{reich_collaborative_2019}.
Other classical machine-learning methods like Gaussian Processes~\cite{zimmer2020influenza},
Generalized Linear models~\cite{chakraborty_what_2018} and nearest-neighbor-based regression~\cite{chakraborty2014forecasting} have also been adapted.

Recent works have also used deep learning-based methods that are flexible to various data sources and capture complex temporal patterns.
While some use off-the-shelf recurrent neural models~\cite{venna2018novel},  others exploit
important characteristics of epidemic dynamics such as dynamically modeling sequence similarity across seasons~\cite{adhikari2019epideep}
and uncertainty with past seasons~\cite{kamarthi2021doubt}, exploiting spatial relations~\cite{deng2020cola,kamarthi2022camul}
as well as leveraging priors from traditional mechanistic models~\cite{rodriguez2022einns,Gao2021STANSA}.
However, most previous works train only from past data for epidemics they forecast and do not leverage useful background
knowledge from a large amount of epidemic data of other diseases collected in the past.

\section{Addditional details on model architecture and training}
\label{sec:addntrain}
We use a 6-layer transformer encoder with 8 attention heads each for \model.
For all pre-training and all downstream tasks, we set the segment size $P=4$ and $\gamma$ as 0.2 for \randmask and 0.1 for \lastmask tasks.
We use learning rate of $10^{-4}$ for pre-training on all SSL tasks and during training simultaneously and use early stopping for training, training to a maximum of 5000 epochs.
We found that pre-training for up to 5000 epochs on all SSL tasks simultaneously was sufficient, as longer pre-training did not improve SSL-related losses or downstream performance significantly. During training, we set 5000 epochs as the maximum, but we observed that most downstream tasks required 1500-2500 epochs to converge and reach the early stopping criteria. Since the datasets in most tasks could fit into the GPU, we set the batch size to be equal to the number of training data points. 

The models were trained on Nvidia Tesla V100 GPU.
We also provide a link to anonymized code and datasets\footnote{Anonymized code link: \url{https://anonymous.4open.science/r/EmbedTS-3F5D/}}.

\section{Training time and memory}
\label{sec:timemem}
We compare the average training time till convergence and memory used by \model and baselines in Table \ref{tab:timemem}.
We observe that the training time and memory consumption of \model is similar to neural baselines while providing significantly more accurate forecasts. Note that
FUNNEL and EB are non-deel learning statistical models that use lower parameters and hence use significantly less training time and memory but provide worse performance.

\begin{table}[h]
\centering
\caption{Average training time and maximum memory taken by each of the baselines and \model for each disease.}
\label{tab:timemem}
\scalebox{0.9}{
\begin{tabular}{c|cccc|cccc}
                & \multicolumn{4}{c}{Average Training time(min)}   & \multicolumn{4}{c}{Max. Memory(GB)}              \\
Model/Benchmark & Flu-US & Flu-Japan & Crypto. & Typhoid & Flu-US & Flu-Japan & Crypto. & Typhoid \\ \hline
AF              & 37.9   & 31.6      & 29.7              & 49.5    & 4.2    & 3.8       & 4.9               & 3.7     \\
IF              & 31.6   & 42.5      & 35.9              & 55.1    & 4.5    & 3.7       & 4.3               & 3.2     \\
PT              & 46.7   & 41.3      & 44.8              & 41.2    & 4.7    & 3.5       & 4.8               & 4.1     \\
DL              & 32.5   & 31.7      & 31.6              & 47.2    & 3.2    & 3.7       & 3.7               & 3.5     \\
TN              & 42.7   & 37.5      & 39.1              & 51.7    & 4.3    & 4.7       & 4.2               & 4.3     \\
MICN            & 36.3   & 39.2      & 36.4              & 48.1    & 3.1    & 3.2       & 3.7               & 3.2     \\ \hline
EF              & 27.4   & 22.5      & 29.3              & 47.2    & 2.8    & 2.1       & 3.5               & 3.1     \\
ED              & 39.1   & 42.7      & 39.6              & 53.6    & 3.2    & 2.7       & 3.4               & 3.1     \\
EB              & 3.4    & 3.2       & 3.9               & 3.5     & 0.1    & 0.1       & 0.1               & 0.1     \\
FUNNEL          & 0.6    & 0.5       & 0.9               & 0.2     & 0.1    & 0.1       & 0.13              & 0.1     \\ \hline
PEM             & 35.4   & 25.5      & 29.2              & 64.5    & 4.7    & 3.5       & 4.8               & 4.1    
\end{tabular}
}
\end{table}

Further, we measure the average training time taken by \model to match the forecast RMSE of the baselines in Table \ref{tab:timecomp}.
We observe that \model matches previous state-of-art performance in much less training time before beating it when trained to convergence.

\begin{table}[h]
\centering
\caption{Comparison of training time taken by \model to match the performance of the best-performing baseline for each benchmark.}
\label{tab:timecomp}
\begin{tabular}{c|cccc}
                                                                                                             & Flu-US & Flu-Japan & Cryptosporidiodia & Typhoid \\ \hline
\begin{tabular}[c]{@{}c@{}}Avg. training time taken\\  to reach performance \\ of best baseline\end{tabular} & 19.5   & 20.2      & 23.7              & 25.9    \\
\begin{tabular}[c]{@{}c@{}}TIme taken by \\ best baseline\end{tabular}                                       & 27.4   & 22.5      & 29.3              & 48.1   
\end{tabular}
\end{table}

\begin{figure}[htbp]
  \centering
  \begin{subfigure}{0.45\textwidth}
    \includegraphics[width=\linewidth]{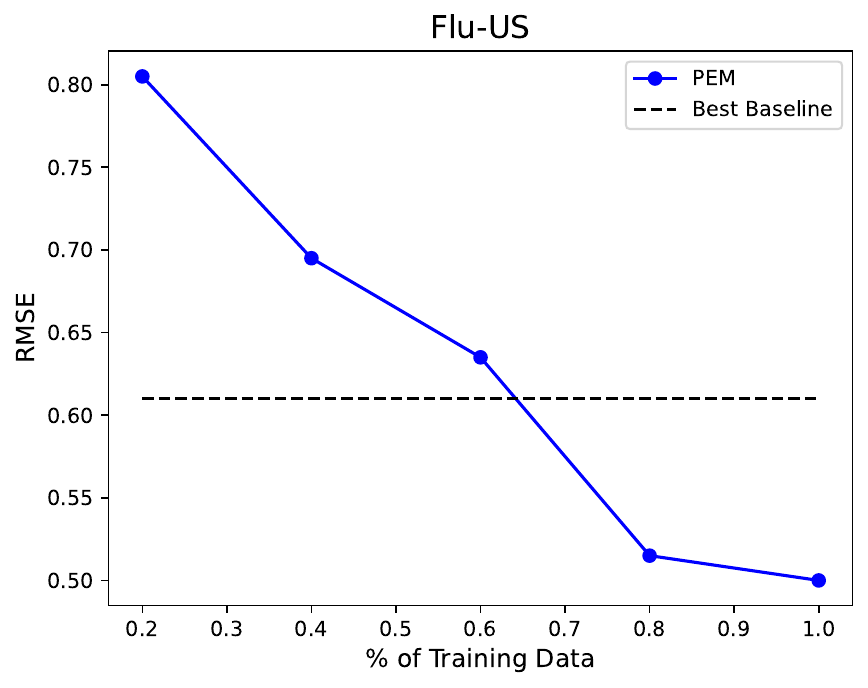}
    \caption{Flu-US}
  \end{subfigure}
  \hfill
  \begin{subfigure}{0.45\textwidth}
    \includegraphics[width=\linewidth]{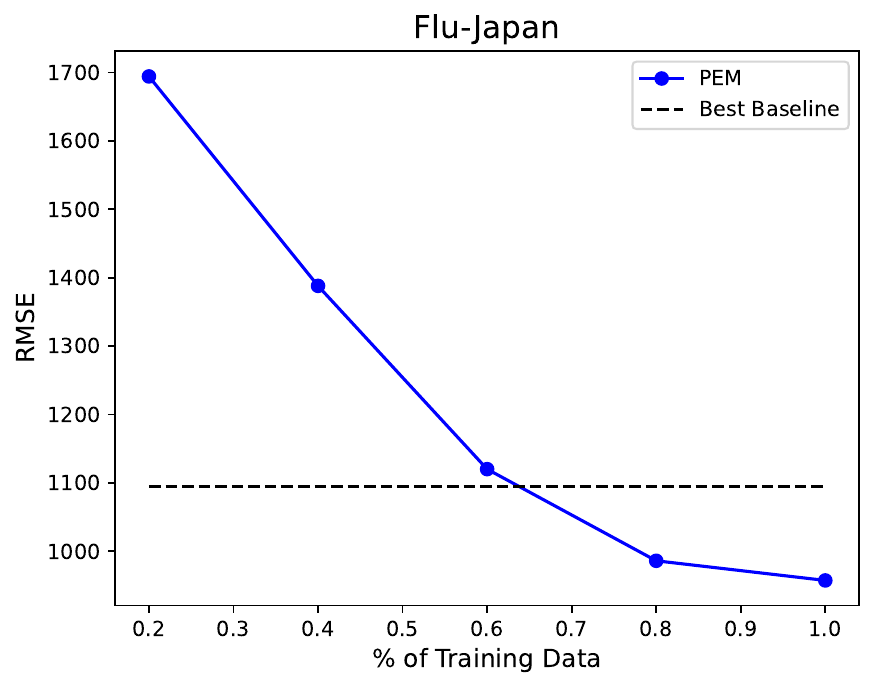}
    \caption{Flu-Japan}
  \end{subfigure}
  
  \vspace{1cm} % Space between the top and bottom figures
  
  \begin{subfigure}{0.45\textwidth}
    \includegraphics[width=\linewidth]{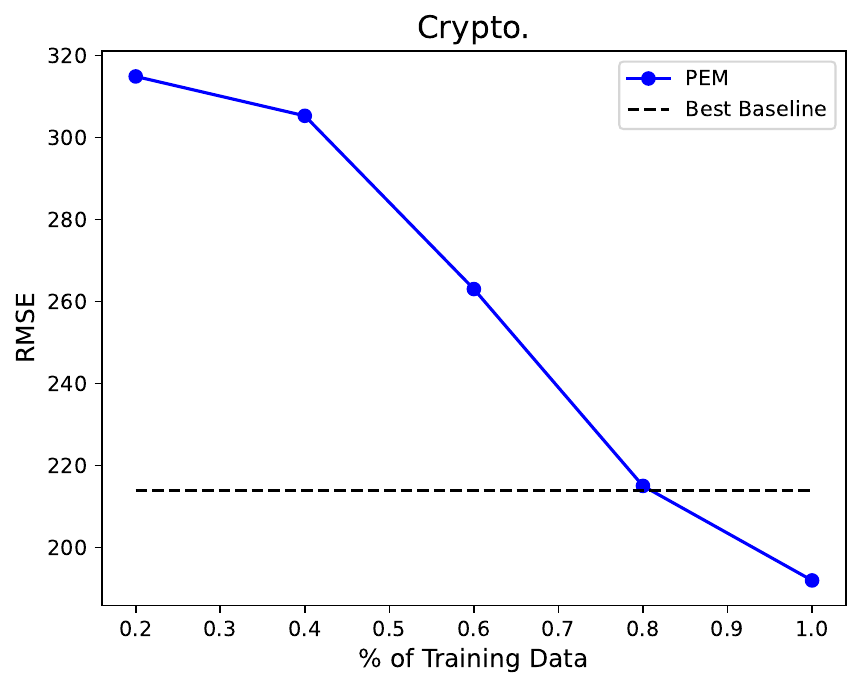}
    \caption{Crypto.}
  \end{subfigure}
  \hfill
  \begin{subfigure}{0.45\textwidth}
    \includegraphics[width=\linewidth]{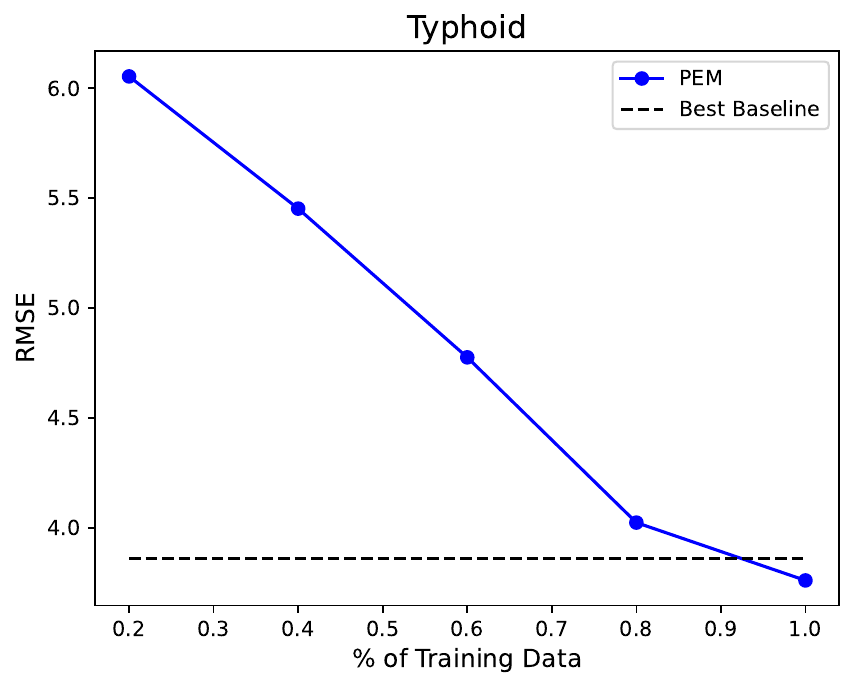}
    \caption{Typhoid}
    \label{fig:perf_percent}
  \end{subfigure}
  
  \caption{Performance of \model with varying fractions of training data. Performance in averaged over 5 runs. Note that in most cases \model's performance is superior to best baseline using less than 80\% of data.}
\end{figure}

\section{Adapting to unseen diseases during pre-training (Q2)}
\label{sec:adapt}
One of the important goals of pre-training on a large number of multi-domain
disease datasets is to capture underlying patterns and information that are observed across time-series of
multiple diseases that can be generalized to newer training datasets as well as previously unseen diseases during pre-training.
The diseases considered in Section \ref{sec:perf} had past data used during pre-training.
In this section, we evaluate how well \model adapts to scenarios where the disease of the training dataset is not used during pre-training.

\paragraph{Forecasting on unseen diseases}
\begin{table}[h]
    \centering
    \caption{Comparison of forecasting performance (RMSE) of \model removing the disease used for training for pre-training with the original \model
        and performance of the best baseline.}
    \label{tab:exclude}
    \scalebox{0.95}{
        \begin{tabular}{c|ccc}
            Dataset          & \multicolumn{1}{l}{Best Baseline} & \multicolumn{1}{l}{PEM} & \multicolumn{1}{l}{PEM-ExcludeTrain} \\\hline
Influenza-US     & 0.62                              & 0.5                     & 0.61                                 \\
Influenza-Japan  & 1466                              & 957.2                   & 997.6                                \\
Cryptosporidosis & 214                               & 192                     & 217.8                                \\
Typhoid          & 4.67                              & 3.76                    & 4.58                                

        \end{tabular}
    }
\end{table}

For each of the training tasks, we pre-train \model removing the disease used for training from $\datapre$.
We call this version of \model as \model-ExcludeTrain.
We compare \model-ExcludeTrain with \model and baselines in Table \ref{tab:exclude}.
While \model-ExcludeTrain's performance is worse compared to \model, in most cases its performance is comparable to if not better than the best baseline for each of the
forecasting tasks.

\paragraph{Case-study on Covid-19}
\begin{table*}[h]
    \centering
    \caption{Forecasting performance on the previously unseen Covid-19 mortality in US from June 2020 to June 2021.}
    \label{tab:covid}
    \begin{tabular}{c|cccccc|ccc|c}
Week ahead & AF         & IF            & PT         & DL            & TN         & MICN & EF         & ED   & EB   & PEM           \\ \hline
1          & 36.3       & \textbf{25.2} & 31.6       & \textbf{26.1} & {\ul 29.3} & 27.4 & 32.7       & 48.2 & 45.2 & 29.7          \\
2          & 44.5       & \textbf{37.1} & 42.7       & 42.4          & 44.7       & 41.5 &  38.9 & 53.2 & 49.7 & {\ul 38.4} \\
3          & { 59.3} & 69.2          & {\ul 55.2} & { 56.9}    & 59.1       & 54.7 & 53.7       & 79.3 & 73.4 & \textbf{48.6} \\
4          & 66.2       & 84.7          & {\ul 59.1} & 59.2          & 63.3       & 59.1 & 68.2       & 81.4 & 85.9 & \textbf{52.6} \\
Avg        & 51.6       & 54.1          & 47.2       & 46.2          & 49.1       & {\ul 45.7} & 48.4       & 65.5 & 63.6 & \textbf{42.3}
\end{tabular}
\end{table*}
We further provide a realistic case study to illustrate the importance of adapting to unseen diseases from pre-training by evaluating the performance of \model
and baselines on the novel Covid-19 pandemic.
We focus on forecasting weekly mortality from Covid-19 in the US~\cite{cramer2022evaluation}.
We do not use any Covid-19 related data in $\datapre$ and only use past Covid-19 data for training \model for each prediction week via the real-time forecasting setup similar to Section \ref{sec:perf}.
The results are summarized in Table \ref{tab:covid}.
On average, we observe a 2\% improvement in forecasting performance over the best baseline with respectable 4\% and 12\% improvement in harder three and four-week
ahead forecasts.
Therefore, \model can successfully leverage pre-training to adapt to even unseen novel pandemics like Covid-19.

\section{Ablation Studies (Q4)}
\label{sec:ablation}
In this section, we study the impact of various model design choices on the performance of \model as well as the parameter sensitivity of some important hyperparameters of \model.

\paragraph{Importance of segmentation and reversible instance normalization}

The superior performance of \model is the result of various design choices related to model architecture as well as pre-training methods.
We studied the impact of each of the SSL tasks in Section \ref{sec:sslresults}.
Here, we observe the impact of important architectural choices of \model on top of the transformer architecture: using segmentation and instance normalization~\cite{kim2021reversible}.
Segments of input time-series are used as tokens instead of individual time-stamps to provide a better semantic representation of the temporal locality of the time-series.
We use reversible instance normalization to accommodate time-series of various magnitudes as well as provide robustness against the distributional shift in individual time-series data.

\begin{table}[h]
    \centering
    \caption{Ablation study of the impact of SSL, segmentation and normalization on \model performance.}
    \label{tab:ablation}
    \scalebox{0.9}{
        \begin{tabular}{c|c|ccc}
            Task                            & Disease           & \multicolumn{1}{l}{PEM-No Segments} & \multicolumn{1}{l}{PEM-No Reversible Norm.} & \multicolumn{1}{l}{PEM} \\ \hline
\multirow{4}{*}{Forecasting}    & Flu-US       & 0.96                                & 0.54                                   & \textbf{0.5}            \\
                                & Flu-Japan   & 1373.7                              & 10165                                  & \textbf{957.2}          \\
                                & Crypto. & 257.2                               & 229.4                                  & \textbf{192}            \\
                                & Typhoid           & 4.81                                & 4.16                                   & \textbf{3.76}           \\ \hline
\multirow{2}{*}{Peak week}      & Flu-US       & 7.26                                & 5.39                                   & \textbf{5.18}           \\
                                & Flu-Japan   & 6.33                                & 6.39                                   & \textbf{4.72}           \\ \hline
\multirow{2}{*}{Peak intensity} & Flu-US       & 0.81                                & 0.95                                   & \textbf{0.72}           \\
                                & Flu-Japan   & 1197                                & 1083                                   & \textbf{864}           

        \end{tabular}
    }
\end{table}
The ablation study is summarized in Table \ref{tab:ablation}.
First, we observe that \model with both components performs better than its ablation variants.
We also observe that without segmentation, the performance decreases by about 75\% in forecasting, 35\% in peak week prediction and 27\% in peak intensity prediction,
underperforming many baselines.
Finally, using reversible instance normalization has the most impact on peak intensity prediction at 31\% whereas only decreases forecasting performance by about 8\%.
Therefore, reversible instance normalization helps adapt to and model data around the peaks which can cause distributional shifts in time-series.

\begin{table}[h]
    \centering
    \caption{Influence of important hyperparameters on average forecasting performance. The default hyperparameter values are {\ul underlined}.}
    \label{tab:sensitivity}
    \scalebox{0.9}{
        \begin{tabular}{c|c|cccc}
            Hyperparameter                  & \multicolumn{1}{l}{Value} & \multicolumn{1}{l}{Flu-US} & \multicolumn{1}{l}{Flu-Japan} & \multicolumn{1}{l}{Cryptosporidiosia} & \multicolumn{1}{l}{Typhoid} \\ \hline
\multirow{3}{*}{Segment size}   & 2                         & 0.79                       & 1366.8                        & 247.4                                 & 5.77                        \\
                                & \underline{4}                & \textbf{0.5}               & \textbf{957.2}                & \textbf{192}                          & \textbf{3.76}               \\
                                & 8                         & 0.59                       & 996.2                         & 229.8                                 & 4.69                        \\ \hline
\multirow{3}{*}{\randmask $\gamma$} & 0.1                       & 0.55                       & 973.7                         & 219.5                                 & 4.13                        \\
                                & \underline{0.2}              & \textbf{0.5}               & \textbf{957.2}                & \textbf{192}                          & \textbf{3.76}               \\
                                & 0.4                       & 0.62                       & 1079.5                        & 286.9                                 & 6.05                        \\ \hline
\multirow{3}{*}{\lastmask $\gamma$} & \underline{0.1}              & \textbf{0.5}               & \textbf{957.2}                & 192                                   & 3.76                        \\
                                & 0.2                       & 0.53                       & 1026.8                        & \textbf{186.3}                        & \textbf{3.51}               \\
                                & 0.4                       & 0.68                       & 1277.5                        & 287.2                                 & 5.37 
        \end{tabular}
    }
\end{table}
\paragraph{Hyperparameter sensitivity analysis}
We also study important hyperparameters of \model on performance in downstream forecasting tasks.
We vary the length of the segments $P$ of the input time-series as well as tune the hardness of the SSL tasks \randmask and \lastmask
by tuning the value of $\gamma$ for each of the tasks.
The average forecasting performance is summarized in Table~\ref{tab:sensitivity}.
We observe that the default hyperparameters of the segment size ($P=4$), $\gamma=0.2$ for \randmask and $\gamma=0.1$ for \lastmask generally perform the best if not close to best across multiple diseases.
Therefore, the important \textit{hyperparameters are not sensitive to specific downstream tasks}.
We also observe that increasing $\gamma$ to a higher value of 0.4 quickly degrades the performance in general since
the reconstruction task gets increasingly harder with an increase in $\gamma$.
%\bibliography{references}
%\bibliographystyle{iclr2024_conference}

\end{document}